\documentclass[sigconf]{acmart}

\usepackage{balance}
\usepackage{bm}
\usepackage{multirow}
\usepackage{caption,subcaption}

\usepackage[capitalize]{cleveref}
\crefname{section}{Sec.}{Secs.}
\Crefname{section}{Section}{Sections}
\Crefname{table}{Table}{Tables}
\crefname{table}{Tab.}{Tabs.}

\AtBeginDocument{%
  }

\copyrightyear{2024}
\acmYear{2024}
\setcopyright{rightsretained}
\acmConference[ICMR '24]{Proceedings of the 2024 International Conference on Multimedia Retrieval}{June 10--14, 2024}{Phuket, Thailand}
\acmBooktitle{Proceedings of the 2024 International Conference on Multimedia Retrieval (ICMR '24), June 10--14, 2024, Phuket, Thailand}
\acmDOI{10.1145/3652583.3658088}
\acmISBN{979-8-4007-0619-6/24/06}
\begin{document}

\title{Improving Video Corpus Moment Retrieval with Partial Relevance Enhancement}

\author{Danyang Hou}
\affiliation{%
  \institution{CAS Key Laboratory of AI Safety, Institute of Computing Technology, Chinese Academy of Sciences}
  \institution{University of Chinese Academy of Sciences}
  \city{Beijing}
  \country{China}}
\email{houdanyang18b@ict.ac.cn}

\author{Liang Pang}
\authornote{Corresponding author}
\affiliation{%
  \institution{CAS Key Laboratory of AI Safety, Institute of Computing Technology, Chinese Academy of Sciences}
  \institution{University of Chinese Academy of Sciences}
  \city{Beijing}
  \country{China}}
\email{pangliang@ict.ac.cn}

\author{Huawei Shen}
\affiliation{%
  \institution{CAS Key Laboratory of AI Safety, Institute of Computing Technology, Chinese Academy of Sciences}
  \institution{University of Chinese Academy of Sciences}
  \city{Beijing}
  \country{China}}
\email{shenhuawei@ict.ac.cn}

\author{Xueqi Cheng}
\affiliation{%
  \institution{CAS Key Laboratory of AI Safety, Institute of Computing Technology, Chinese Academy of Sciences}
  \institution{University of Chinese Academy of Sciences}
  \city{Beijing}
  \country{China}}
\email{cxq@ict.ac.cn}

\renewcommand{\shortauthors}{Danyang Hou, Liang Pang, Huawei Shen, and Xueqi Cheng}

\begin{abstract}
Video Corpus Moment Retrieval (VCMR) is a new video retrieval task aimed at retrieving a relevant moment from a large corpus of untrimmed videos using a text query. 
The relevance between the video and query is partial, mainly evident in two aspects:~(1)~Scope: The untrimmed video contains many frames, but not all are relevant to the query. Strong relevance is typically observed only within the relevant moment.~(2)~Modality: The relevance of the query varies with different modalities. Action descriptions align more with visual elements, while character conversations are more related to textual information.
Existing methods often treat all video contents equally, leading to sub-optimal moment retrieval. We argue that effectively capturing the partial relevance between the query and video is essential for the VCMR task.
To this end, we propose a Partial Relevance Enhanced Model~(PREM) to improve VCMR. VCMR involves two sub-tasks: video retrieval and moment localization. To align with their distinct objectives, we implement specialized partial relevance enhancement strategies. For video retrieval, we introduce a multi-modal collaborative video retriever, generating different query representations for the two modalities by modality-specific pooling, ensuring a more effective match. For moment localization, we propose the focus-then-fuse moment localizer, utilizing modality-specific gates to capture essential content. We also introduce relevant content-enhanced training methods for both retriever and localizer to enhance the ability of model to capture relevant content.  
Experimental results on TVR and DiDeMo datasets show that the proposed model outperforms the baselines, achieving a new state-of-the-art of VCMR. The code is available at \url{https://github.com/hdy007007/PREM}.

\end{abstract}




\maketitle


\begin{figure}
    \begin{center}
    \includegraphics[width=0.46\textwidth]{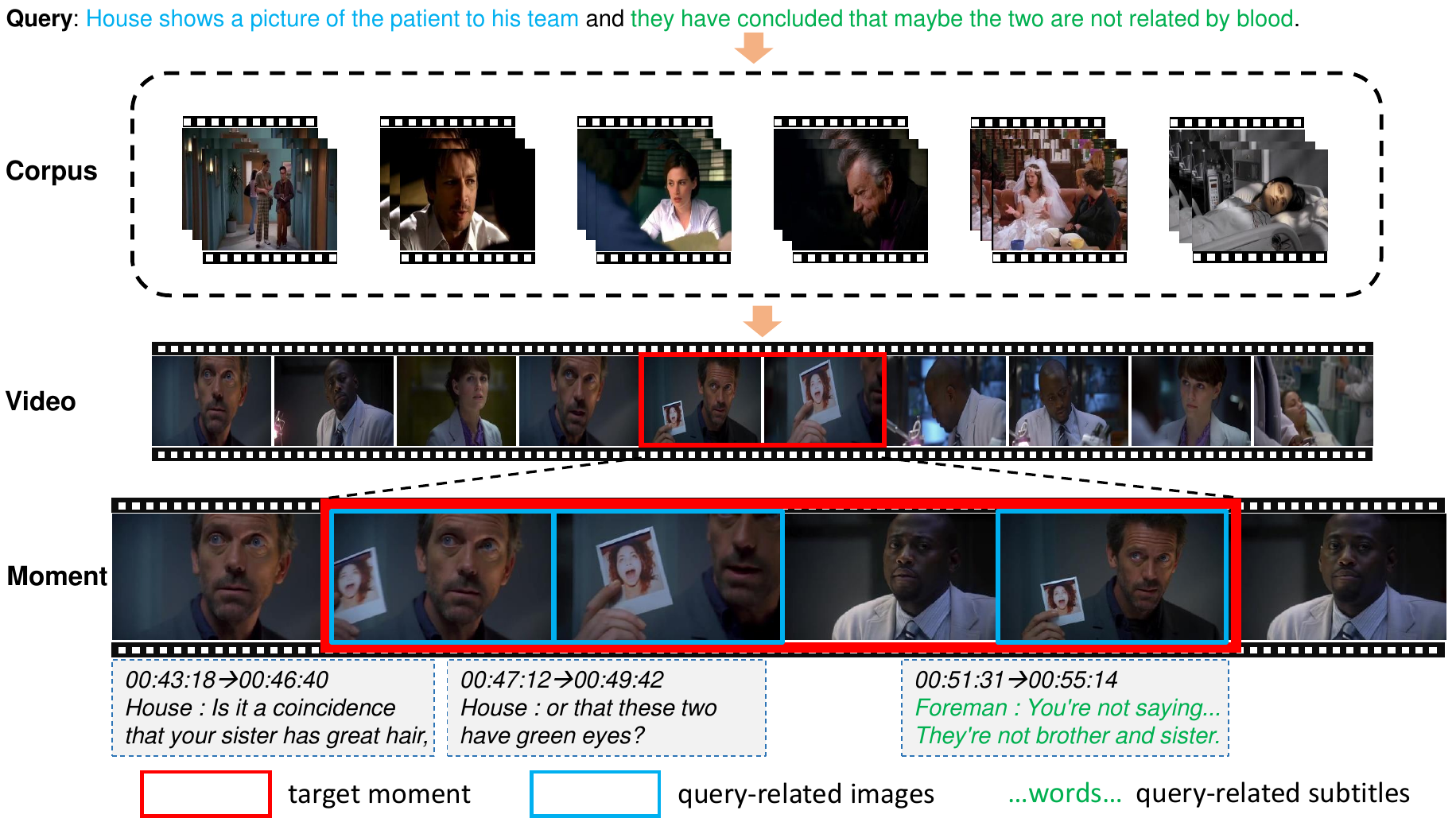}
    \end{center}
    \caption{In VCMR task, only a small part of the untrimmed video is explicitly related to the query, i.e., the content within the target moment. And the relevance of query to information from different modalities in the video varies. }
    \label{fig:teaser}
\end{figure}
\section{Introduction}

The rise of video-sharing applications has led to a dramatic increase in videos on the Internet. With a large corpus of videos, it is vital to help users find the information they need efficiently. Many video search engines can receive text queries from users to retrieve videos, but the retrieved videos are usually the original full videos in the video corpus. If users are looking for fine-grained information, such as a classic shot from a movie or a key step in a cooking video, these search engines can hardly meet the demand.
We have the opportunity to address this challenge thanks to the recently proposed Video Corpus Moment Retrieval~(VCMR)~\cite{temporal, lei2020tvr} task that requires retrieving a video moment via a natural language query from a collection of untrimmed videos, where the moment is a temporal segment of a video. The task can be decomposed into two sub-tasks:  Video Retrieval~(VR), which aims to retrieve the relevant video that potentially contains the target moment using a text query, and Single Video Moment Retrieval~(SVMR), which requires localizing the target moment semantically related to the query in the retrieved video. 

The uniqueness of the VCMR task, compared to the typical text-video retrieval~\cite{gabeur2020multi}, stems from the information richness inherent in video. Unlike typical text-video retrieval where the video is precisely trimmed to align entirely with the text query, VCMR involves untrimmed video that contains a wealth of information, with only a small fraction being directly relevant to the query. The partial relevance between the query and video is evident in two key aspects:~(1)~ \textbf{Scope of the relevant content}: Not all video content is related to the query, and explicitly relevant content is often limited to the context of the target moment, as illustrated in~\Cref{fig:teaser}. In the TVR~\cite{lei2020tvr} dataset, the average duration of the relevant moment for a query constitutes only 11\% of the average total video duration. Therefore, it is essential to enhance the capability of retrieval model of capturing these limited yet significant relevant content in the video.~(2)~\textbf{Discrepancy in relevance across modalities}: The video contains multi-modal information, such as visual information and textual information~(subtitle), each modality exhibiting distinct query relevance. For instance, "House shows his team a picture of the patient" of the query in~\Cref{fig:teaser} predominantly pertains to visual information, depicting an action. Conversely, "they have concluded that maybe the two are not related by blood" leans more towards the textual modality, involving a conversation between characters. As Wang et al.~\cite{wang2022modality} emphasize, recognizing the varying query relevance across different modalities is crucial for better multi-modal alignment in multi-modal retrieval.
For these reasons, we argue that capturing the content of both the scope and modality in the video relevant to the query can help moment retrieval.

However, the existing methods for VCMR treat content across all positions and modalities in the video equally. This treatment yields a model incapable of prioritizing relevant content, resulting in sub-optimal moment retrieval. To address this, we propose a Partial Relevance Enhanced Model~(PREM)~for the VCMR task. 
We employ distinct partial relevance enhancement strategies for the two sub-tasks to align with their respective objectives. Specially, for VR, we introduce a multi-modal collaborative video retriever that contains a modality-specific pooling component. This component generates different query representations for each modality, ensuring a comprehensive match between the query and the content of the two modalities in the video. For SVMR, we propose a focus-then-fuse moment localizer. In this localizer, features from the two modalities are fed to  modality-specific gates to capture query-relevant elements. Following this, a fine-grained multi-modal fusion is employed for accurate moment localization. To enhance the emphasis of the model on content in the relevant scope to query, we introduce relevant content-enhanced contrastive learning and adversarial training for the two modules respectively.

We evaluate the proposed model on two benchmarks, TVR~\cite{lei2020tvr} and DiDeMo~\cite{anne2017localizing}. The results show the effectiveness of PREM, achieving new state-of-the-art results.

Our contributions are as follows:
\begin{itemize}
    \item We propose a partial relevance enhanced model for the VCMR task, which encourages the model to capture query-related content within the untrimmed video.
    \item We introduce a multi-modal collaborative video retriever for VR and a focus-then-fuse moment localizer for SVMR, ensuring a thorough match between the query and the two modalities of video. And two relevant content-enhanced training objectives are employed to emphasize content in query-relevant scope of video.
    \item The experimental results on TVR~\cite{lei2020tvr} and DiDeMo~\cite{anne2017localizing} show that the proposed model outperforms other baselines, achieving new state-of-the-art results.
\end{itemize}

\section{Related Work}
We first provide a brief overview of works related to the two sub-tasks of VCMR, followed by a detailed introduction to recent works for VCMR.

\noindent \textbf{Text-video retrieval} is a typical cross-modal retrieval task which aims to retrieve releated videos from a corpus based on a text query. While similar to video retrieval in the VCMR task, the distinction lies in the fact that in text-video retrieval, most of the video content is relevant to the query, whereas, in VCMR, only a small segment is relevant to the query.  Works for text-video retrieval generally fall into two categories based on the interaction mode between the query and video, namely late fusion and early fusion.
Late-fusion methods~\cite{song2019polysemous, dong2021dual, patrick2020support, yang2020tree, gabeur2020multi} employ separate encoders for text and video to map them into a shared semantic space. These models exhibit high efficiency when video representations are pre-computed and indexed offline, as only query representation and the similarity between query and videos need to be computed during inference.
On the other hand, early-fusion methods~\cite{chen2020fine, wang2021t2vlad, jin2021hierarchical, wu2021hanet, song2021spatial, han2021fine} adopt fine-grained cross-modal interactions using attention mechanism~\cite{Bahdanau_attention, vaswani2017attention}, enhancing retrieval accuracy. However, these methods face a trade-off between retrieval efficiency and accuracy as the efficiency is constrained by the necessity to perform the entire online computations of query-video relevance.

\noindent \textbf{Temporal language grounding} is a task similar to SVMR, which requires localizing a moment relevant to a given text query from a video.  Temporal language grounding can be seen as a special case of VCMR, with only one video in the corpus for each query.   According to the way of predicting moment, the existing works for temporal language grounding can be divided into proposal-based and proposal-free. 
The proposal-based method~\cite{liu2018cross,xu2019multilevel,chen2019semantic,xiao-etal-2021-natural,chen2018temporally,zhang2019man,zhang2021multi,liu2021context} begins by generating several candidate proposals. These candidates are then scored based on their alignment with the query, with the proposal exhibiting the strongest alignment being chosen as the final answer.
The proposal-free method~\cite{yuan2019find,chen2020rethinking,zeng2020dense,li2021proposal,ghosh2019excl,chen2019localizing,zhang2020span}, predicts the start and end times of the moment directly, bypassing the need to pre-extract proposals as candidates.
Recently, several studies~\cite{lei2021detecting, moon2023query, liu2022umt, cao2021pursuit} introduce DETR\cite{carion2020end}-based grounding models, which streamline the complex post-processing steps of earlier methods.
It is impractical to apply a temporal language grounding model to predict moments across all videos for VCMR, given the immense computations. A video retrieval module is needed to narrow down the videos to a very small set.

\noindent \textbf{Video corpus moment retrieval} is proposed by Escorcia et al.~\cite{temporal}, and then Lei et al.~\cite{lei2020tvr} introduce a new dataset TVR~\cite{lei2020tvr} specifically for VCMR. 
Existing methods for VCMR fall into two categories based on how they address the learning of the two sub-tasks, namely one-stage~\cite{lei2020tvr, zhang2021video, li2020hero, zhang2020hierarchical, yoon2022selective} and two-stage methods~\cite{hou2021conquer, zhang2023video, chen2023cross}. The one-stage method treats VCMR as a multi-task learning problem, employing a shared backbone with two distinct heads to simultaneously learn VR and SVMR tasks.
One-stage methods can be further categorized into late-fusion methods and early-fusion methods, similar to the taxonomy in text-video retrieval task. XML~\cite{lei2020tvr}, ReLoCLNet~\cite{zhang2021video} and HERO~\cite{li2020hero} are late-fusion models. XML is the first model proposed for the VCMR task, and ReLoCLNet enhances the performance of the late-fusion model by contrastive learning. HERO is a video-language pre-trained model, which significantly improves retrieval accuracy.  HAMMER~\cite{zhang2020hierarchical} and SQuiDNet~\cite{yoon2022selective} belong to the early-fusion method. HAMMER employs a hierarchical attention mechanism to make deep interactions between query and video. SQuiDNet leverages causal reasoning to reduce bad retrieval bias for video retrieval.
On the other hand, the two-stage approach leverages specialized modules for each sub-task, combining the benefits of both late-fusion and early-fusion methods. It employs the late-fusion model as the video retrieval module for fast video retrieval and utilizes the early-fusion model as the moment localization module for accurate moment localization. CONQUER~\cite{hou2021conquer}, DFMAT~\cite{zhang2023video} and CKCN~\cite{chen2023cross} are two-stage models. The three models use the video retrieval head of the trained HERO as video retriever and propose early-fusion moment localizers. CONQUER proposes a moment localizer based on context-query attention~(CQA)~\cite{yu2018qanet}. DFMAT introduces a moment localizer with a multi-scale deformable attention module for multi-granularity feature fusion, while CKCN proposes a calibration network to improve cross-modal interaction. 
Our proposed PREM falls into the two-stage method. Unlike other works for VCMR that treat all locations and modalities in the untrimmed video equally, our model prioritizes content related to the query by modality-specific modules and relevant content-enhanced training.




\section{Method}
In this section, we introduce our proposed model, beginning with an explanation of the VCMR task formulation. We then describe feature extraction. Subsequently, we provide insights into the video retrieval module and moment localization module. Finally, we detail the training and inference processes of model.

\subsection{Task Formulation}
\begin{figure}
    \begin{center}
    \includegraphics[width=0.5\textwidth]{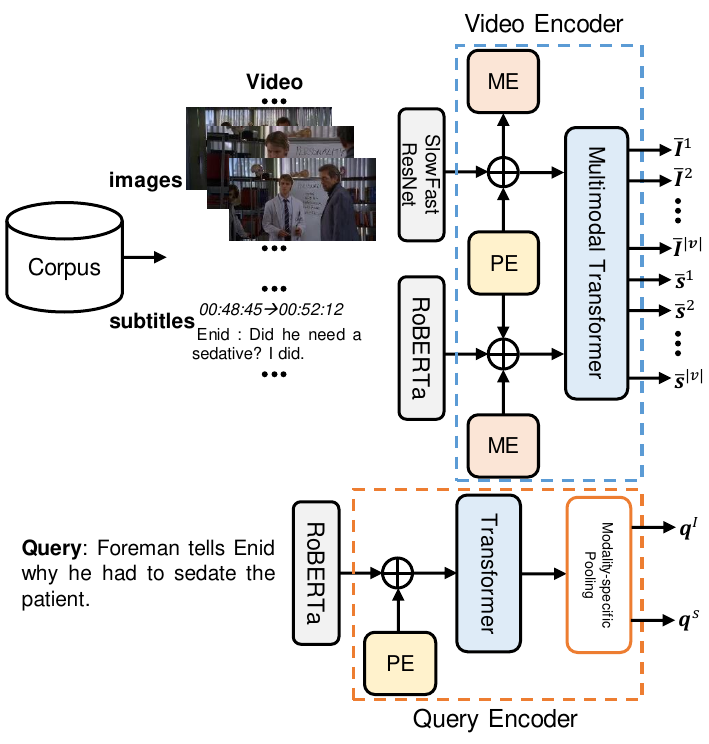}
    \end{center}
    \caption{The video retriever consists of two encoders, a video encoder and a query encoder. 'ME' and 'PE' represent modality and positional embedding, respectively. }
    \label{fig:retriever}
\end{figure}

Given a corpus of videos $\mathcal{V} = \{v_1, v_2,...,v_{|\mathcal{V}|}\}$ where $|\mathcal{V}|$ is the number of videos in corpus and $v_i=\{c^1_i,c^2_i,...,c^{|v_i|}_i\}$ is the $i$-th video which contains $|v_i|$ non-overlapping clips, the goal of VCMR task is to retrieve the most relevant moment $m_*$ from $\mathcal{V}$ using a text query $q=\{w^1,w^2,...,w^{|q|}\}$: 
\begin{equation}
\label{equ:goal}
m_* = \mathop{\rm argmax}\limits_{m} P(m|q, \mathcal{V}),
\end{equation}
where the moment is a temporal segment $[\tau^{st}, \tau^{ed}]$ in video $v_*$. VCMR can be decomposed to two sub-tasks, VR and SVMR. The goal of VR is to retrieve the video that contains the target moment:
\begin{equation}
\label{equ:goal_vr}
v_* = \mathop{\rm argmax}\limits_{v} P(v|q, \mathcal{V}).
\end{equation}
And SVMR aims to localize moment from the retrieved video:
\begin{equation}
\label{equ:goal_svmr}
m_* = \mathop{\rm argmax}\limits_{m} P(m|v_*,q).
\end{equation}
The prediction of the target moment depends on the probabilities of start and end positions:
\begin{equation}
\label{equ:st_ed_prob}
P(m|v_*,q) = P(\tau^{st}|v_*,q)\cdot P(\tau^{ed}|v_*,q). 
\end{equation}
We use a video retriever to model $P(v|q, \mathcal{V})$ and a moment localizer to model $P(m|v_*,q)$ shown in~\Cref{fig:retriever} and~\Cref{fig:localizer} respectively. 

\subsection{Feature Extraction}
We leverage the pre-trained networks to extract the initial features of input for our model. For the query feature, we utilize a pre-trained RoBERTa~\cite{liu2019roberta} to extract the token feature $\bm{w}^j$ of each word in a query. For video features, we extract clip features of images and subtitles.  In particular, a clip feature of the image is taken by max-pooling the image features in a short duration~(1.5 seconds), which is extracted by pre-trained SlowFast~\cite{feichtenhofer2019slowfast} and ResNet~\cite{he2016deep}~(concatenate the two features as image feature) in a clip, and a clip feature of subtitle is taken by max-pooling the token features extracted by RoBERTa in a clip. Thus, a clip consists of an image feature and a subtitle feature $\bm{c}^j = \{\bm{I}^j, \bm{s}^j\}$, if the clip contains no subtitle, $\bm{s}^j$ is set to a vector of zeros. The features are mapped to a semantic space $\mathbb{R}^D$ by fully-connect layers. In the following paper, we use bold notations to denote vectors.

\subsection{Multi-modal Collaborative Video Retriever}
\label{sec:retrieval}
The goal of the video retriever is to select top-$K$ videos $\mathcal{V}^*$ from the corpus $\mathcal{V}$ given the query $q$ based on the relevance, where $K$ is much smaller than $|\mathcal{V}|$. 
The retriever in our proposed model is a late-fusion architecture comprising two encoders: one for query and another for video. To capture the relevance between the query and the different modalities within the video, we introduce modality-specific pooling, which produces distinct query representations for each modality. Furthermore, we utilize relevant content-enhanced contrastive learning to improve capability of retriever to capture content within the query-related scope of the video.

\noindent \textbf{Query Encoder}
The token features of a query with positional embeddings are fed to a one-layer Transformer to output $\bar{\bm{w}}^j$. Because the words semantically match to different modalities may be different shown in~\Cref{fig:teaser}, we employ modal-specific pooling to generate two modality-specific query representations for query-image relevance and query-subtitle relevance, denoted as $\bm{q}_I$ and $\bm{q}_s$. Specifically, we first calculate the likelihood of each word belonging to a modality, then determine the weights based on the likelihood, and finally conduct a weighted summation of the word representations as the modality-specific representation:
\begin{equation} 
o^j = \bm{W}_d \bar{\bm{w}}^j   , \\
\ \alpha^j = \frac{{\rm exp}(o^j)}{\sum \limits_{i=1}^{|q|} {\rm exp}(o^i)}   , \\
\ \bm{q}^d =  \sum \limits_{j=1}^{|q|} \alpha^j \bar{\bm{w}}^j  ,
\end{equation}
where $\bm{W}_d \in \mathbb{R}^{D \times 1}$ is a fully-connect layer which outputs a scalar $o^j$ to represent the likelihood, $d \in \{I, s\}$ for visual modality and textual modality. $\alpha^j$ is softmax normalized weight of $j$-th word. And $\bm{q}^d$ is a modal-specific representation for vision or text.

\noindent \textbf{Video Encoder}
For the $i$-th video, we feed subtitle features and image features with positional embeddings and modality embeddings to a one-layer multi-modal Transformer that simultaneously captures intra-modal and inter-modal dependencies to output each contextual image representation $\bar{\bm{I}}_i^j$ and subtitle representation $\bar{\bm{s}}_i^j$.

Since only part of the content in the video is related to the query, the similarity score $S^R_i$ between the query and $i$-th video is the average of max-pooling of query-image similarities and max-pooling of query-subtitle similarities. We use the cosine similarity as the similarity score ${\rm sim}()$:
\begin{equation}
\begin{aligned}
{\rm sim}(\bm{q}^d, \bar{\bm{d}}_i^j) &= {\rm cos} (\bm{q}^d , \bar{\bm{d}}_i^j), \ \ d\in\{I, s\}  \\ 
\phi_d  &= \mathop{\rm max}\limits_{1 \leq j \leq |v_i|} {\rm sim}(\bm{q}^d, \bar{\bm{d}}_i^j),  \\
S^R_i &= \frac{\phi_I + \phi_s}{2}.
\end{aligned}
\end{equation}

\begin{figure}
    \begin{center}
    \includegraphics[width=0.48\textwidth]{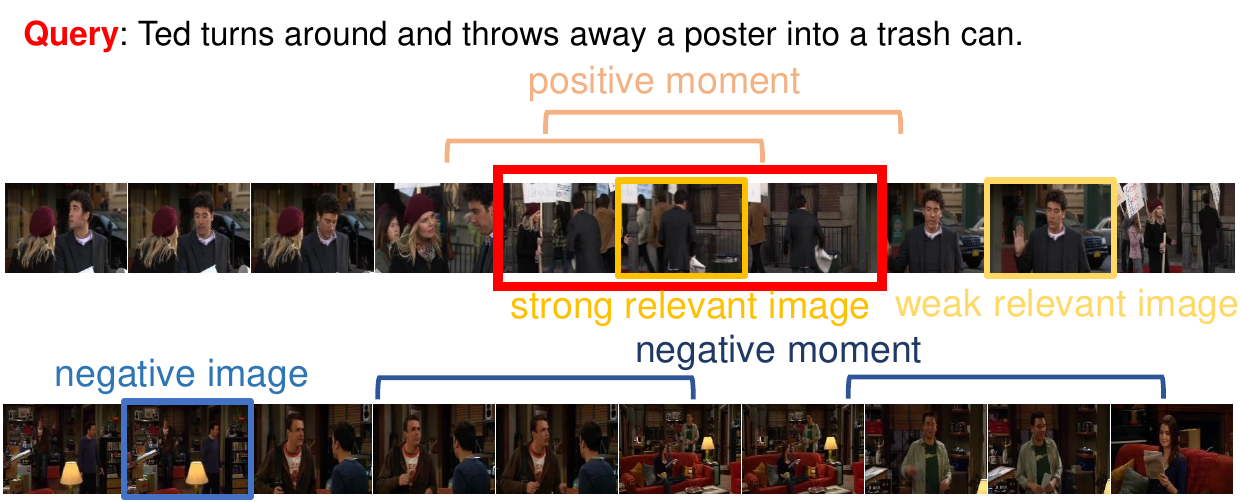}
    \end{center}
    \caption{Relevant and negative images sampling for contrastive learning of video retriever, and positive and negative moments sampling for adversarial learning of moment localizer. The segment in the video with a red border is the query-related moment.}
    \label{fig:samples}
\end{figure}

\noindent \textbf{Relevant Content-enhanced Contrastive Learning} Contrastive learning~\cite{tian2020contrastive} is widely leveraged to train late-fusion retrieval models, such as~\cite{karpukhin2020dense}. However, there are few studies on retrieval that query partially related to the target. To this end, we propose a relevant content-enhanced contrastive learning method for training our video retriever to put more focus on query-relevant content. The essence of our approach lies in selecting content relevant to the query as a positive sample and increasing the similarity between the query representation and the query-related content representation computed by the retriever. As depicted in~\Cref{fig:teaser}, the content explicitly related to the query is always within the relevant moment. Therefore, we select the image and subtitle with the highest similarity to the query within the moment shown in~\Cref{fig:samples} as strong relevant samples, the overall similarity denoted as $S^R_{++}$. For negative samples, we choose the images and subtitles with the highest similarity to the query in each negative video, the similarity is denoted as $S^R_{-}$. We employ InfoNCE~\cite{van2018representation} loss to train retriever:
\begin{equation}
\label{equ:loss_retriever}
\mathcal{L}^v_{++} =  -log \frac{{\rm exp}(S^R_{++} / t)}{{\rm exp}(S^R_{++} / t) + \sum\limits_{i = 1}^{n} {\rm exp}(S^R_{-,i}/t)}, 
\end{equation}
where $t$ is temperature set to 0.01, and $n$ is the number of negatives. We adopt in-batch negative sampling strategy that all videos in the batch except positive video can be used for negative sampling. 

\begin{figure*}
    \begin{center}
    \includegraphics[width=0.88\textwidth]{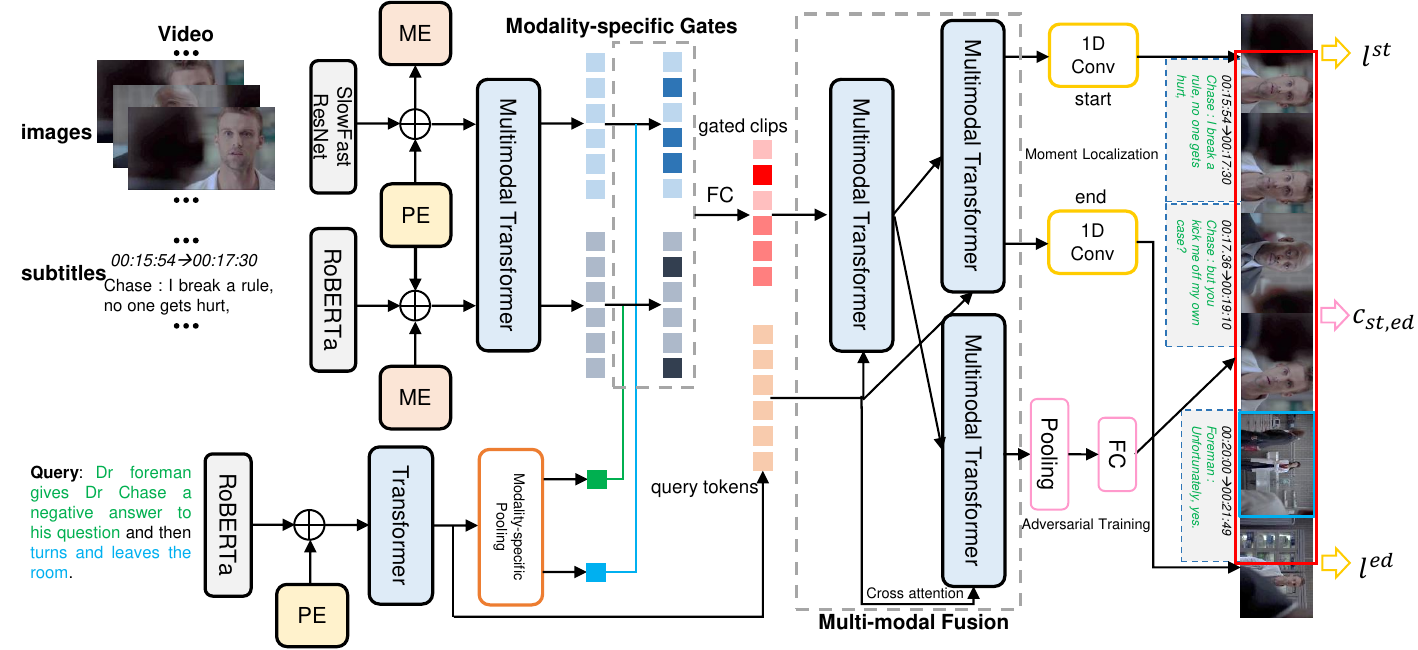}
    \end{center}
    \caption{Moment localizer contains two key components, modality-specific gates and a multi-modal fusion network to align query and multi-modal content in the video. We use the square to represent a vector. }
    \label{fig:localizer}
\end{figure*}

Beyond considering relevant content within the moment, we also explore relevant content in the video that falls outside the moment, as it may still exhibit relevance to the query, such as adjacent images or subtitles to the moment, which we call weak relevant content. The weak relevant content is the image and subtitle with the highest similarity to the query but not within the moment shown in~\Cref{fig:samples}, and the similarity is denoted as $S^R_+$. The loss for weak relevant content $\mathcal{L}^v_{+}$ is the same as that in~\Cref{equ:loss_retriever}. The  query-to-video loss is:
\begin{equation}
\mathcal{L}^v = \mathcal{L}^v_{++}  + \lambda * \mathcal{L}^v_{+}, 
\label{equ:loss_strong_weak}
\end{equation}
where $\lambda$ is a  hyper-parameter set to 0.5. 

We also incorporate video-to-query loss $\mathcal{L}^q$~(only using the strong positive sample), as in most works on cross-modal retrieval. The sum of $\mathcal{L}^v$ and $\mathcal{L}^q$ is the overall loss for the video retriever.

\subsection{Focus-then-fuse Moment Localizer}

Moment localizer shown in~\Cref{fig:localizer} uses the query to localize the target moment $m_*$ in the top-$K$ retrieved videos $\mathcal{V}_*$. The proposed localizer is based on early-fusion architecture to explore deeper interactions between query and video for accurate moment localization. 
Because the retrieved videos are narrowed down to a small range, the amount of computations is acceptable.
The localizer comprises encoders, modality-specific gates focusing on significant content from different modalities, and a multi-modal Transformer for query and video fusion. In addition to the regular loss for training the localizer (optimizing the prediction for moment boundaries), we propose an additional relevant content-enhanced adversarial loss to enhance the partial relevance.


\noindent \textbf{Encoders}
Similar to the video retriever, we obtain multi-modal contextual representations of images $\bar{\bm{I}}$ and subtitles $\bar{\bm{s}}$ in the video using a multi-modal Transformer and specific pooling to generate two representations $\bm{q}^I$ and $\bm{q}^s$ for different modalities. We aslo retain word representations for token-to-token interactions between the query and video.

\noindent \textbf{Modality-specific Gates}
Because the query-related parts of modalities are different shown in~\Cref{fig:teaser}, we refer to~\cite{wang2017gated} to design modality-specific gates to capture query-related content of images and subtitles:
\begin{equation} 
\bm{\hat{d}}^j = {\rm norm}((\hat{\bm{W}}_d\ \bar{\bm{d}}^j)  \odot  \bm{q}^d) \odot \bar{\bm{d}}^j    , d \in \{I ,s\} 
\end{equation}
where $\hat{\bm{W}}_d \in \mathbb{R}^{D \times D}$ is a fully-connect layer, and $\bm{\hat{d}}^j$ is the $j$-th gated image representation or subtitle representation. $\bm{q}^d$ is modality-specific query representation for image or subtitle. ${\rm norm}$ is L2-normalization. $\odot$ is element-wise product.

We then fuse the gated representations of two modalities in a clip by a fully-connect layer:
$\bm{\hat{c}}^j =  {\rm FC}([\bm{\hat{I}}^j;\bm{\hat{s}}^j])$ , 
where $[;]$ is concatenation and $\hat{\bm{c}}^j$ is gated representation  of the $j$-th clip. The gated video representations is $\bm{\hat{v_i}} = \{\bm{\hat{c}}^1_i, ..., \bm{\hat{c}}^{|v_i|}_i \}$.

\noindent \textbf{Multi-modal Fusion} We leverage multi-modal Transformer to fuse token representations in query and clip representations for deep cross-modal interactions. Specifically, we input the gated clip representations to a two-layer Transformer. In each layer, alongside the self-attention layer for clip interactions, there is an additional cross-attention layer capturing fine-grained cross-modal dependencies between the video and the query.

To predict the start and end times of the target moment, we employ two 1D-convolution networks to capture dependencies between adjacent clips and output boundary scores $l^{st}_i$ and $l^{ed}_i$, indicating whether the $i$-th clip serves as a boundary for the moment. The training objective is to make the score of the clip that is the correct boundary of the moment higher than the other clips. To ensure comparability of predicted moments across all retrieved videos, we adopt shared normalization (Shared-Norm)~\cite{clark-gardner-2018-simple}, a common practice in Open-domain QA systems. In addition to the clip scores predicted by the query and relevant video, we input the query and irrelevant videos to the localizer to predict clip scores, which serve as additional negative samples for training. The training is based on cross entropy loss:
\begin{equation}
\mathcal{L}^{st} = -log   \frac{{\rm exp}(l^{st}_{+})}{\sum \limits_{a=1} \limits^{n+1} \sum \limits_{b=1} \limits^{|v_a|} {\rm exp}(l^{st}_{a,b}) }  , 
\mathcal{L}^{ed} = -log   \frac{{\rm exp}(l^{ed}_{+})}{\sum \limits_{a=1} \limits^{n+1} \sum \limits_{b=1} \limits^{|v_a|} {\rm exp}(l^{ed}_{a,b}) } , 
\end{equation}
where $l^{st}_{+}$ is start score of the ground truth moment, and $l^{st}_{a,b}$ is start score of the $b$-th clip in video $v_a$. 

\noindent \textbf{Relevant Content-enhanced Adversarial Training} Like the content-enhanced contrastive learning for the video retriever, we use an adversarial learning method which is classification task to improve the localizer’s focus on query-relevant content.
The positive samples for classification are obtained by sampling all segments in the video with an IoU greater than 0.7 with the ground truth moment, as shown in~\Cref{fig:samples}. 
The negatives consist of the top-5 predicted moments in each negative video, a dynamic set. The features for classification are derived from another Transformer as depicted in~\Cref{fig:localizer}. Specifically, the feature for classification is obtained through max-pooling Transformer representations of positive or negative moments. The classification network is a fully-connect network that outputs a score $c_{i, j}$, representing whether the segment within [i, j] is positive or negative. The loss $\mathcal{L}^{c}$ for adversarial training is based on binary cross entropy loss. The overall loss for training moment localizer is:
\begin{equation}
\mathcal{L} = \mathcal{L}^{st} + \mathcal{L}^{ed} + \gamma * \mathcal{L}^{c},
\end{equation}
where $\gamma$ is a hyper-parameter set to 0.8.

\subsection{Training and Inference}
We implement a stage-wise training strategy, initially training the video retriever using text-video pairs. Subsequently, the trained video retriever is employed to sample negative videos as hard negatives for Shared-Norm to train the moment localizer.

In the inference phase, we begin by retrieving the top-10 videos using the video retriever for a given query. Subsequently, we employ the moment localizer to localize the moment within the selected 10 videos. Notably, moment classification in adversarial training does not contribute to moment prediction. The score of the predicted moment relies on both the video retrieval score and boundary scores:
\begin{equation}
S = \frac{S^R}{t} + l^{st} + l^{ed},
\end{equation}
where $t$ is the temperature in contrastive learning. We use $S$ to rank predicted moments in the retrieved videos.


\begin{table}[]
\caption{VR results on the TVR validation set and DiDeMo testing set. $\dag$: HERO without pre-training on a large dataset. $*$: HERO fine-tuned using our relevant content-enhanced contrastive learning.}
\begin{tabular}{llcccc}
\hline
Dataset                 & Model     & R@1   & R@5   & R@10  & R@100 \\
\hline
\multirow{6}{*}{TVR}    & XML~\cite{lei2020tvr}       & 18.52 & 41.36 & 53.15 & 89.59 \\
                        & ReLoCLNet~\cite{zhang2021video} & 22.63 & 46.54 & 57.91 & 90.65 \\
                        & ${\rm HERO}^{\dag}$~\cite{li2020hero}      & 19.44 & 42.08 & 52.34 & 84.94 \\
                        & HERO~\cite{li2020hero}      & 29.01 & 52.82 & 63.07 & 89.91 \\
                        \cline{2-6} 
                        & PREM      & 26.24 & 51.35 & 63.08 & \textbf{92.50} \\
                        & ${\rm HERO}^{*}$      & \textbf{32.88} & \textbf{55.62} & \textbf{65.35} & 91.26 \\
\hline
\multirow{5}{*}{DiDeMo} & XML~\cite{lei2020tvr}       & 6.23  & 19.35 & 29.95 & 74.16 \\
                        & ReLoCLNet~\cite{zhang2021video} & 5.53  & 18.25 & 27.96 & 71.42 \\
                        & ${\rm HERO}^{\dag}$~\cite{li2020hero}      & 5.11  & 16.35 & 33.11 & 68.38 \\
                        & HERO~\cite{li2020hero}      & \textbf{8.46}  & 23.43 & \textbf{34.86} & 75.36 \\
                        \cline{2-6} 
                        & PREM      & 8.03  & \textbf{23.68} & 34.59 & \textbf{75.75} \\
\hline
\end{tabular}
\label{tab:vr_results}
\end{table}

\begin{table}[]
\caption{Ablation of video retriever on TVR validation set. 'Text': subtitles provided by TVR dataset. 'Vision': images in video. 'SR': strong relevant positive sample. 'WR': weak relevant positive sample.}
\begin{tabular}{cccccccc}
\hline
Text & Vision & SR & WR & R@1   & R@5   & R@10  & R@100 \\
\hline
 $\surd$ &  $\surd$ & $\surd$   &  $\surd$  & \textbf{26.24} & \textbf{51.35} & \textbf{63.08} & \textbf{92.50} \\
  & $\surd$  & $\surd$   &  $\surd$  & 15.12 & 36.69 & 47.99 & 86.09 \\
$\surd$  & &  $\surd$  &  $\surd$  & 16.02 & 33.43 & 43.08 & 80.64 \\
$\surd$  & $\surd$  &  $\surd$  &    & 25.76 & 50.89 & 62.94 & 92.33 \\
$\surd$  & $\surd$  &    &    & 23.64 & 47.99 & 59.29 & 91.08 \\
\hline
\end{tabular}
\label{tab:ablation_vr}
\end{table}

\section{Experiments}
In this section, we begin by introducing datasets and metrics, followed by a discussion of implementation details. Subsequently, we present experimental results and compare them with baselines. Next, we conduct ablation studies to analyze the components of model. Finally, we provide visualizations and case studies.

\subsection{Datasets and Evaluation Metrics}
\label{sec:dataset}
We conduct experiments on two datasets: \textbf{TVR}\cite{lei2020tvr} and \textbf{DiDeMo}\cite{anne2017localizing}. TVR is built on TV Shows whose video consists of images and subtitles, and contains 17435, 2179, and 1089 videos on the training, validation, and testing sets. The average length of the videos is 76.2 seconds, while the average length of the moments is 9.1 secs. There are 5 queries whose average length is 13.4 for each video. DiDeMo is built on YFCC100M\cite{thomee2015new}, a video dataset whose videos are from the real world, with only images and no subtitles in the video. DiDeMo contains 8395, 1065, and 1004 videos on training, validation, and testing, respectively. Most of the videos are about 30 seconds, and the average duration of moments is 6.5 seconds. Each video contains 4 queries whose average length is 8.

We follow the metrics in \cite{lei2020tvr} as evaluation metrics. For the VCMR task, the evaluation metric is \textbf{R@$K$, IoU=$p$} that represents the percentage of at least one predicted moment whose Intersection over Union(IoU) with the ground truth exceeds $p$ in the top-$K$ retrieved moments. The two sub-tasks are also evaluated. The metric of the SVMR task is the same as that of the VR task, but the evaluation is conducted in only ground truth video for each query. As for the VR task, the metric is  \textbf{R@$K$}, which denotes the percentage of predictions that the correct video in the top-$K$ ranked videos.

\begin{figure}
    \begin{center}
    \includegraphics[width=0.30\textwidth]{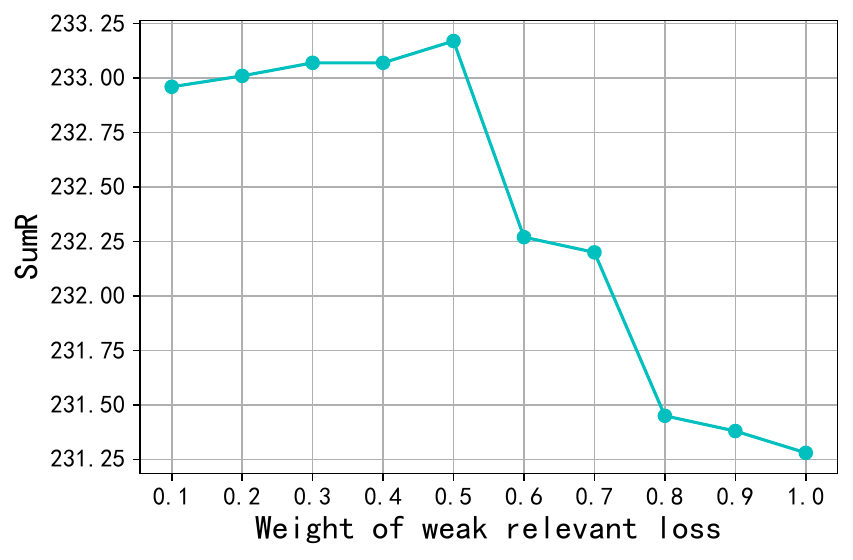}
    \end{center}
    \caption{Effect of the weight of weak relevant loss on VR. SumR is sum of R@K (K = 1, 5, 10 , 100).}
    \label{fig:weight_weak_relevant}
\end{figure}

\begin{table}[]
\caption{VR results on TVR validation set using different pooling methods.}
\begin{tabular}{lcccc}
\hline
Pooling          & R@1   & R@5   & R@10  & R@100 \\
\hline
Mean             & 23.74 & 48.20 & 59.48 & 91.43 \\
Max              & 25.19 & 49.48 & 61.17 & 91.94 \\
Modality-specific & \textbf{26.24} & \textbf{51.35} & \textbf{63.08} & \textbf{92.50} \\
\hline
\end{tabular}
\label{tab:pooling_vr}
\end{table}

\begin{table}[]
\caption{SVMR and VCMR results (R@1, IoU=0.5, 0.7) on TVR validation set. $*$: early-fusion one-stage models. $^{\bigstar}$: late-fusion one-stage models. $\dag$: two-stage models that use HERO as the video retriever.}
\begin{tabular}{lcccc}
\hline
\multirow{2}{*}{Model} & \multicolumn{2}{c}{SVMR} & \multicolumn{2}{c}{VCMR} \\
                       & 0.5         & 0.7        & 0.5         & 0.7        \\
\hline
${\rm HAMMER}^{*}$~\cite{zhang2020hierarchical} (Arxiv'20) & - & - & 9.19 & 5.13 \\ 
${\rm SQuiDNet^{*}}$~\cite{yoon2022selective} (ECCV'22) & 41.31 & 24.74 & - & 8.52 \\ 
${\rm XML}^{\bigstar}$~\cite{lei2020tvr} (ECCV'20) & 29.56 & 13.05 & 6.17 & 2.91 \\
${\rm ReLoCLNet}^{\bigstar}$~\cite{zhang2021video} & 31.65 & 14.80 & 7.99 & 4.11 \\
${\rm HERO}^{\bigstar}$~\cite{li2020hero} (EMNLP'20) & 32.22 & 15.30 & - & 5.13 \\
${\rm CONQUER}^{\dag}$~\cite{hou2021conquer} (MM'21) & 42.24 & 22.84 & 13.57  & 7.76 \\
${\rm DMFAT}^{\dag}$~\cite{zhang2023video} (TCSVT'23) & 43.51 & 23.26 & 13.75 & 7.99 \\
${\rm CKCN}^{\dag}$~\cite{chen2023cross} (TMM'23) & 43.38 & 23.18 & 13.69 & 7.92 \\
\hline
PREM    & \textbf{46.29} & \textbf{24.83} & 16.08 & 9.01 \\
${\rm PREM}^{\dag}$   & 46.21 & 24.71 & \textbf{19.74} & \textbf{11.12} \\
\hline
\end{tabular}
\label{tab:vcmr_tvr}
\end{table}

\begin{table}[]
\caption{SVMR and VCMR results (R@1, IoU=0.5, 0.7) on DiDeMo testing set.}
\begin{tabular}{lcccc}
\hline
\multirow{2}{*}{Model} & \multicolumn{2}{c}{SVMR} & \multicolumn{2}{c}{VCMR} \\
                       & 0.5         & 0.7        & 0.5         & 0.7        \\
\hline
XML~\cite{lei2020tvr} (ECCV'20)                   & -           & -          & 2.36        & 1.59       \\
ReLoCLNet~\cite{zhang2021video} (SIGIR'21)             & 34.81       & 26.71      & 2.28        & 1.71       \\
HERO~\cite{li2020hero} (EMNLP'20)          & 39.20       & 30.19      & 3.42        & 2.79       \\
CONQUER~\cite{hou2021conquer} (MM'21)                & 38.17       & 29.90       & 3.31        & 2.79       \\
DMFAT~\cite{zhang2023video} (TCSVT'23)                 & -           & -          & 3.44        & 2.89       \\
CKCN~\cite{chen2023cross} (TMM'23)                    & 36.54       & 28.99      & 3.22        & 2.69      \\
\hline
PREM                   & \textbf{40.79}       & \textbf{33.77}      & \textbf{3.51}        & \textbf{2.98}       \\
\hline
\end{tabular}
\label{tab:svmr_vcmr_didemo}
\end{table}

\subsection{Implementation Details}

\noindent \textbf{Feature Extraction} For query, we use 768D RoBERTa feature provided by~\cite{lei2020tvr}. For video, the duration of a clip is 1.5 seconds, and the FPS for sampling images is 3. We use 4352D SlowFast+ResNet feature provided by~\cite{li2020hero} as the video feature and use 768D RoBERTa feature provided by~\cite{lei2020tvr} as the subtitle feature.

\noindent \textbf{Architecture} For Transformer, we use the architecture of vanilla Transformer~\cite{vaswani2017attention} with the hidden size 384 and the intermediate size 1536. The number of heads for multi-head attention is 4. 

\noindent \textbf{Training} We train video retriever for 100 epochs with the batch size 256. As for moment localizer, we sample 4 and 2 negative videos for each query from top-100 ranked videos on TVR and DiDeMo respectively, and train it for 15 epochs with the batch size 32. Both the retriever and localizer are trained by AdamW with the learning rate 0.0001 and the weight decay 0.01 in a single NVIDIA Tesla V100 GPU. More details are shown in our released code.


\subsection{Comparison with Baselines}
We compare our proposed PREM with baselines on VCMR task including late-fusion models XML~\cite{lei2020tvr}, ReLoCLNet~\cite{zhang2021video}, and HERO~\cite{li2020hero}, early-fusion models HAMMER~\cite{zhang2020hierarchical}, SQuiDNet~\cite{yoon2022selective}, and two-stage models CONQUER~\cite{hou2021conquer}, DMFAT~\cite{zhang2023video} and CKCN~\cite{chen2023cross}. 

\noindent \textbf{VR} Reported in~\Cref{tab:vr_results}, our proposed PREM outperforms other methods in both benchmarks, with the exception of HERO. HERO benefits from pre-training on two large text-video datasets, TVR~\cite{lei2020tvr} and HowTo100M~\cite{miech2019howto100m}, to acquire additional knowledge. In contrast, our model is exclusively trained on the TVR training set.
HERO without pre-training achieves sub-optimal results. HERO fine-tuned using our proposed relevant content-enhanced contrastive learning substantially improves the retrieval accuracy, showing the effectiveness of the partial relevance enhancement in VR task. ReLoCLNet also employs contrastive learning to train the retrieval model; however, it underperforms our model on both datasets, because it neglects to consider query-relevant content when sampling positive samples. Instead, our sampling incorporates both strong and weak relevant clips to encourage the model to prioritize query-relevant content in the video.

\begin{table}[]
\caption{Ablation of moment localizer on TVR validation set.}
\begin{tabular}{lcccc}
\hline
\multirow{2}{*}{Model}          & \multicolumn{2}{c}{SVMR} & \multicolumn{2}{c}{VCMR} \\
                           & 0.5         & 0.7        & 0.5         & 0.7        \\
\hline
Full Model                 & \textbf{46.29}       & \textbf{24.83}      & \textbf{16.08}       & \textbf{9.01}       \\
\  w/o Text                   & 41.06       & 21.77      & 13.48       & 7.52       \\
\  w/o Image                  & 25.92       & 13.81      & 8.32        & 4.49       \\
\  w/o Modality-specific Gates & 44.53       & 23.97      & 15.14        & 8.53       \\
\  w/o Adversarial Training   & 45.39       & 24.25      & 15.60        & 8.71       \\
\  w/o Shared-Norm            & 43.55       & 23.53      & 13.09        &  7.76     \\
\hline
\end{tabular}
\label{tab:ablation_localizer}
\end{table}

\noindent \textbf{SVMR and VCMR} As shown in~\Cref{tab:vcmr_tvr} and~\Cref{tab:svmr_vcmr_didemo}, our proposed model outperforms other baseline methods on two tasks and two datasets, underscoring the effectiveness of our focus-then-fuse moment localizer. The heightened performance of early-fusion and two-stage models compared to late-fusion models can be attributed to the integration of cross-attention in their localizers, facilitating comprehensive interaction of fine-grained information across modalities. Noteworthy is the significant improvement in moment retrieval accuracy on TVR when utilizing HERO finetuned by our proposed relevant content-contrastive learning as the video retriever.

\subsection{Ablation Study}

\noindent \textbf{Video Retriever}
As reported in~\Cref{tab:ablation_vr}, each component of our video retriever contributes to the performance. Both image and subtitle in the video play important roles in video retrieval.
When sampling positive examples for contrastive learning, it is effective to extract strong relevant samples from the query-related moment, as the content within that scope is explicitly related to the query. Furthermore, selecting weak relevant samples from outside the moment is also effective. However, Retrieval accuracy decreases when we exclude relevant content sampling, where we employ the sampling method from the previous methods~\cite{zhang2021video}, i.e., selecting the clip with the highest similarity to the query from all content of correct video as positive sample. This approach can make it challenging to learn the relevance between the query and the query-related moment.

We further investigate the effect of the weak relevant loss weight $\lambda$ in~\Cref{equ:loss_strong_weak} on VR. As shown in~\Cref{fig:weight_weak_relevant}, the retrieval accuracy is highest when the weight is set to 0.5. However, the accuracy decreases with the weight greater than 0.5, because weak relevant samples are not always explicitly related to the query, and a smaller weight can mitigate the negative impact of their uncertain relevance.

\begin{figure}
\begin{minipage}{0.48\textwidth}
\centering
\includegraphics[width = 7.5 cm]{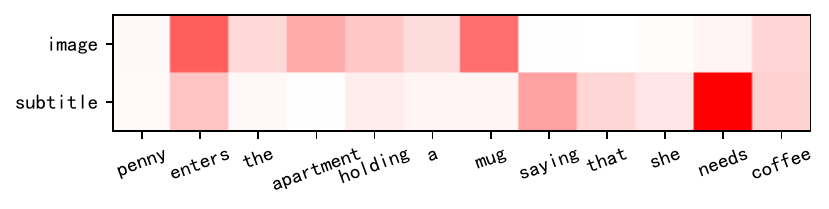}
\subcaption{Heatmap of weights for words to compose modality-specific queries.}
\label{fig:heat_query}
\end{minipage}
\begin{minipage}{0.5\textwidth}  

\centering
\includegraphics[width = 8.3 cm]{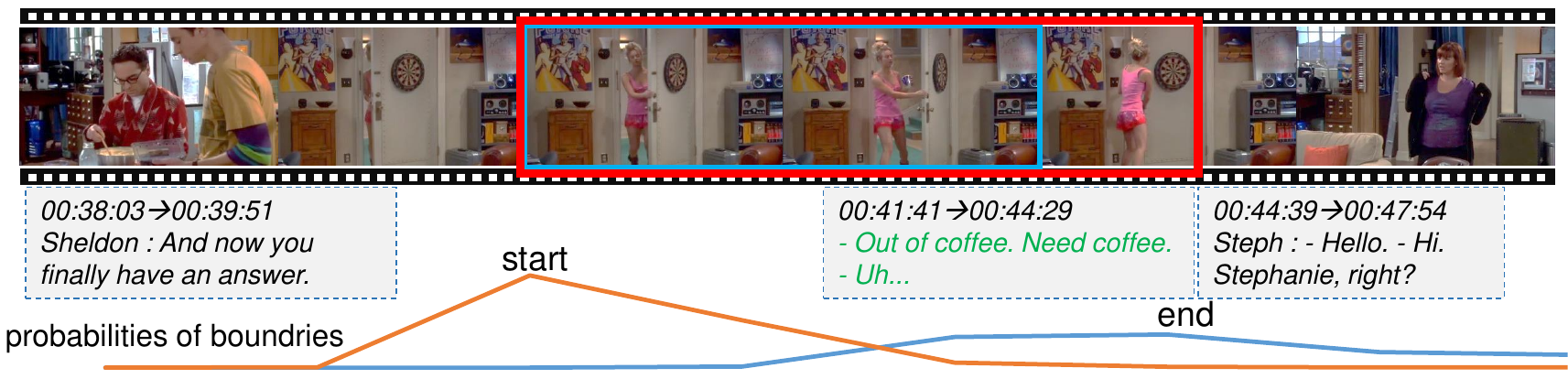}
\subcaption{Confidence scores of start and end boundaries for moment prediction.}
\label{fig:boundaries}
\end{minipage}
\centering
\caption{Visualizations of modality-specific pooling in video retriever and moment prediction of moment localizer. The query is "Penny enters the apartment holding a mug saying that she needs coffee."}
\end{figure}

In~\Cref{tab:pooling_vr}, we report the results of various pooling methods in the retriever. Modality-specific pooling outperforms the two other commonly used pooling methods, confirming the discrepancy in relevance across modalities to the query, as motivated in our approach. A visualization is provided in \Cref{fig:heat_query}, showing the weights of individual words that compose the query representation. Words with higher weights are indicated by a darker shade of red. For instance, “Penny enters the apartment holding a mug” in the query has higher weight for the image query, reflecting its visual nature, whereas “saying that she needs coffee” has higher weight for the subtitle, suggesting a stronger association with textual information.

\noindent \textbf{Moment Localizer}
~\Cref{tab:ablation_localizer} reports the contributions of key components in the moment localizer for the performance on SVMR and VCMR. Unlike the video retriever, the moment localizer places a greater emphasis on visual information than textual information due to the nature of the moment localization task, which requires accurate matching of an action to a query. Both Modality-specific gates and adversarial training contribute to the performance of moment localization, confirming the effectiveness of our partial relevance enhancement strategies. As illustrated in ~\Cref{fig:boundaries}, the localizer utilizes key information from both modalities when predicting the boundaries. The start boundary prediction primarily relies on the image, while the end boundary prediction depends on the subtitle, as the subtitle ‘Need coffee’ is related to ‘saying that she needs coffee’ in the query. Additionally, Shared-Norm plays a crucial role for VCMR by enabling the localizer to localize the target moment across multiple videos.

\subsection{Case Study}

\begin{figure}
\begin{minipage}{0.48\textwidth}
\centering
\includegraphics[width = 8.3 cm]{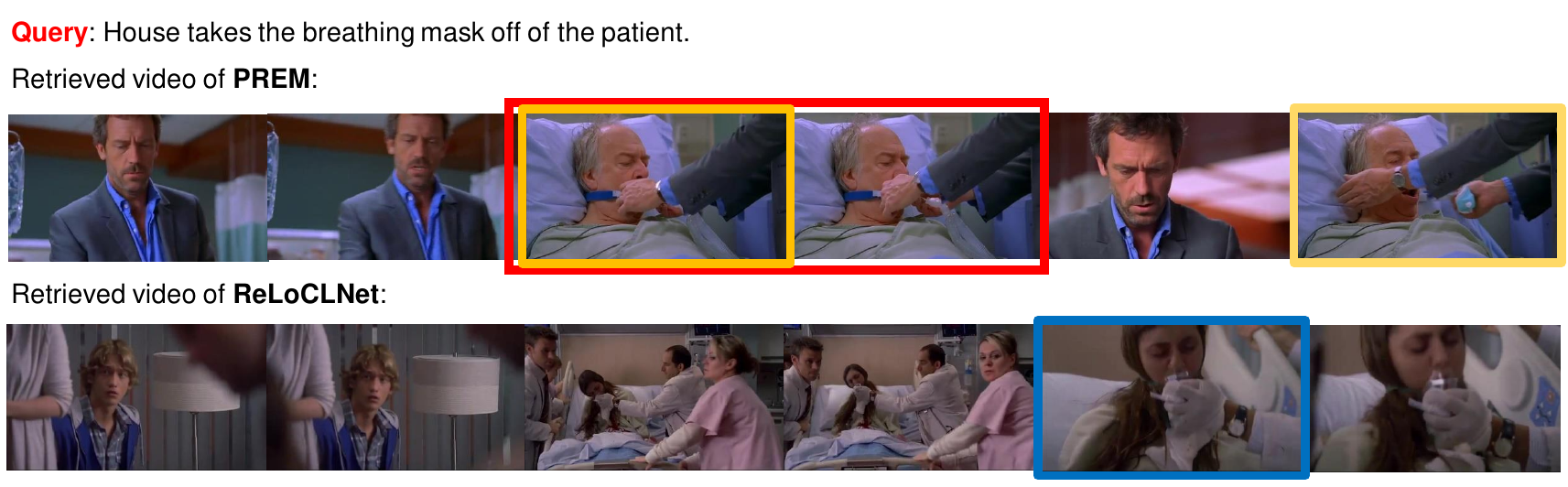}
\subcaption{Video retrieval.}
\label{fig:case_retriever}
\end{minipage}
\begin{minipage}{0.48\textwidth}  
\centering
\includegraphics[width = 8.3 cm]{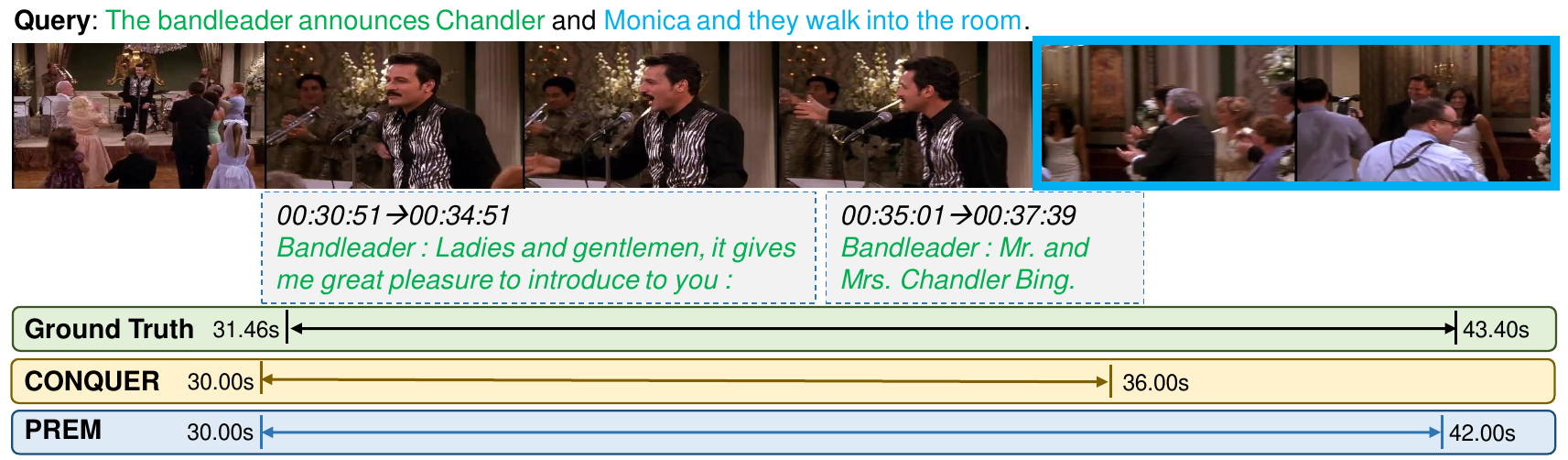}
\subcaption{Single video moment retrieval.}
\label{fig:case_localizer}
\end{minipage}
\centering
\caption{Case study. (a) VR: the images with yellow borders in the video retrieved by PREM are the most similar images to the query within and outside the correct moment, while the image with a blue border in the video retrieved by ReLoCLNet is the most similar to query. (b) SVMR: query-related subtitles or images are distinguished by different colors.}
\label{fig:case}
\end{figure}

We present examples of VR and SVMR in \Cref{fig:case}. In VR, our model ranks the correct video first, and the most similar images to the query are relevant to the query. However, the query and the most similar image in the video retrieved by ReLoCLNet lack semantic alignment; the character is not House, and the action in the video involves putting on a breathing mask rather than taking it off. In SVMR, the moment predicted by the proposed model is closer to the ground truth because it captures the images related to ‘they walk into the room’ to help localize the moment. These two cases validate the effectiveness of our partial relevance enhancement for both VR and SVMR tasks.

\section{Conclusion}

In this paper, we propose a Partial Relevance Enhanced Model (PREM) to improve the VCMR task. For the two sub-tasks of VCMR, we introduce two modules with different partial relevance enhancement strategies: a multi-modal collaborative video retriever for VR and a focus-then-fuse moment localizer for SVMR. To further encourage the model to capture the partial relevance between the query and the video, we propose relevant content-based contrastive learning and adversarial training for the training of the two modules. Extensive experiments on two datasets, TVR and DiDeMo, demonstrate that our proposed model achieves new state-of-the-art results on the VCMR task. The ablation studies and visualizations confirm the effectiveness of the partial relevance enhancement in our proposed model. In the future, we plan to explore additional modalities within the video, such as speech, to enhance the  ability of retrieval model to capture partial relevance.

\section*{ACKNOWLEDGMENTS}
This work was supported by the National Key R\&D Program of China (2022YFB3103700, 2022YFB3103704), the National
Natural Science Foundation of China (NSFC) under Grants No. 62276248 and U21B2046, and the Youth Innovation Promotion Association
CAS under Grants No. 2023111.

\bibliographystyle{ACM-Reference-Format}
\balance
\bibliography{sample-base}

\end{document}